# Temporal transformer CAN encoder with federated lightweight heads for anomaly detection

**Konstantinos Gyftodimos[1*], Kyriakos Chiotis[1], Elena Politi[1], George Dimitrakopoulos[2], and Eirini Liotou[2]**

[1]ANADELTA P.C., Athens, Greece

[2]Department of Informatics and Telematics, Harokopio University of Athens, Greece

[1]{kgyftodimos, politie}@anadelta.hua.gr

[2]{kyriakoschiotis, gdimitra, eliotou}@hua.gr

**Abstract**

Modern vehicles rely on large numbers of Electronic Control Units (ECUs) that constantly exchange information over the Controller Area Network (CAN) bus. Due to the rapidity, structure, and repetition of this communication, even slight variations in timing, payload values, or message patterns can point to unusual activity. Whether due to errors, malfunctions, or deliberate interference, these anomalies are frequently subtle and challenging to identify with conventional methods that handle messages separately or rely on manually created rules. Motivated by this gap, we present a privacy-preserving framework for anomaly detection in in-vehicle networks, based on a Temporal Transformer CAN Encoder with Federated Lightweight Heads, to better capture these irregularities. The detection of subtle temporal and contextual anomalies is made possible by a lightweight Transformer encoder that learns how these signals evolve over time, while a federated learning mechanism enables several vehicles or ECUs to work together to improve a shared model without exchanging raw CAN data. This combination of federated learning and temporal sequence modeling provides robust anomaly detection performance while maintaining efficiency and privacy, according to experiments conducted on open-source datasets.

## 1. Introduction

Controller Area Network (CAN) is the dominant in-vehicle communication protocol, through which the Electronic Control Units (ECUs) of Intelligent Connected Vehicles (ICVs) communicate with each other, facilitating real-time data exchange between critical components. Nevertheless, CAN bus was not designed with inherent, built-in mechanisms that ensure the integrity or correctness of the data being transmitted, with respect in parallel to real-time constraints. Moreover, studies demonstrate that modern vehicles are vulnerable to security attacks such as spoofing, injection, flooding, and timing-based attacks, among others [1]. What is more, in the Software-Defined Vehicle era, where vehicles have been transformed into even more open, connected and powerful systems, the protection of CAN traffic from security or privacy related threats becomes even more critical and essential.

Such attacks or other kinds of non-malicious irregularities, such as faults, degradation or misconfigurations usually appear as disruptions, i.e., anomalies to timing patterns, payload variations or context-aware changes. The detection of such malicious or abnormal CAN behavior requires the employment of models that capture the temporal, contextual and structural dynamics of CAN traffic [2]. The complexity of such solutions is high, since a) CAN messages are short and high-rate, thus studying them in isolation does not easily reveal potential issues, b) temporal structure and timing is often ignored in many Intrusion Detection Systems (IDS), while c) even solutions that capture temporal characteristics are often computationally heavy and require centralized training. However, centralized training entails privacy- and data-sharing risks as well as difficulties in being deployed in resource-constrained ECUs

[3][4][5].

Motivated by these challenges, we propose an anomaly detection method based on a Federated Learning (FL) approach for in vehicle networks. Specifically, our paper introduces a novel framework that processes raw CAN logs, converts them into fixed-length windows, and feeds each window into a Temporal Transformer CAN Encoder, through which temporal representations are extracted. In sequence, our system introduces a lightweight XGBoost anomaly head that is trained to classify window-level embeddings efficiently and thus, predict anomaly distributions per window. Finally, our solution optionally employs FL, as the means to preserve privacy through collaborative learning, as well as efficiency in real, resource-constrained environments.

Our contributions, therefore, can be summarized as follows:

- We introduce a lightweight Transformer model that captures timing behavior of CAN messages. In contrary to most modern IDS systems, our method does not handle each CAN message independently but learns the combined temporal dynamics.
- We employ a lightweight local classifier (XGBoost) as well as FL, so that the proposed solution is highly realistic and suitable for real-time resource-constrained ECUs, respecting in parallel privacy requirements of actual vehicular ecosystems.
- We provide the full pipeline as an end-to-end framework, covering all required aspects and components, from preprocessing to FL.
- We evaluate the proposed framework on real CAN datasets (OTIDS), demonstrating improved detection of timing and payload anomalies.

The remainder of the paper is structured as follows: Section 2 surveys related work in anomaly detection, while Section 3 formalizes the problem and describes the overall framework and training objectives. The proposed Temporal Transformer CAN Encoder and the FL mechanism are discussed in Section 4. Following that, Section 5 describes the experimental setup and evaluation metrics, while Section 6 discusses the achieved results. Lastly, Section 7 concludes the paper.

## 2. Related Work

Driven by recent advances in pattern-learning algorithms and large-scale data analytics, several ML-based techniques have been introduced for detecting anomalies in in-vehicle networks. Such data driven methods utilize large volumes of historical data to identify malicious behaviors in real-time data exchange between critical components. Traditional Machine Learning (ML) algorithms such as Isolation Forest (IF) [6] and One-Class Support Vector Machine (OC-SVM) [7] have been proposed for identifying state changes from features selected with human intervention. To reduce the dependence on prior human feature selection, recent research has increasingly turned to deep learning-based anomaly detection methods.

Thus, recurrent neural networks (RNNs) and long short-term memory networks (LSTMs) have been widely employed to leverage temporal patterns. An LSTM algorithm was proposed in [8] to predict the upcoming frame of the CAN bus and detect attacks by quantifying the deviation between normal and abnormal by comparing the reconstruction error of the input traffic sequence. Similarly, an autoencoder-based approach was proposed to learn latent sequential patterns in CAN data, under the assumption that anomalies would manifest as discrepancies in the encoded representations of IVN sequences compared to those of normal traffic [9]. Liu et al. introduced an Autoencoder Wasserstein Generative Adversarial Network (AE-AnoWGAN) for anomaly detection, using the time-frequency representation (TFR) of vibration signals as input and deriving anomaly scores from the reconstructed TFR. The effectiveness of the approach was demonstrated across multiple public datasets [10].

According to a recent survey by Özdemir et al. [11], many CAN anomaly detection methods still rely on manually constructed, per-message characteristics, ignoring the temporal and structural regularities present in CAN transmission. Dong et al. [12] provide an anomaly detection method based on frequency and arbitration to support these findings, demonstrating that abnormalities in message frequency and arbitration behavior can be reliable indicators of CAN disruptions. Althunayyan et al. [13] provide a hierarchical FL architecture for in-vehicle networks that enables distributed IDS training without revealing raw CAN data. Their results show that FL can enable scalable and cooperative anomaly detection while adhering to strict automotive privacy regulations. Moreover, Ji et al. [14] propose a spatiotemporal anomaly detection framework that employs feature extraction and channel attention to find anomalous patterns in network traffic. Finally, Hanselmann et al. [15] demonstrate how temporal and structural embeddings can effectively capture complex linkages within CAN traffic in their presentation of CANet, an unsupervised deep learning IDS for high-dimensional CAN data.

Our work differentiates from the current state of the art in that, first of all, it combines a lightweight Temporal Transformer encoder with a fast XGBoost classifier, an approach that has not been studied previously. Moreover, the proposed framework has been designed for resource-constrained ECUs, namely in such a way that it increases the efficiency and realism of the solution, on top respecting privacy through the use of FL. Furthermore, our proposal fuses inter-arrival times, payload bytes, and CAN identifiers within a single temporal representation, in contrary to current works that analyze for instance payload or frequency in isolation.

## 3. Proposed Framework

The proposed framework introduces a downstream multi-stage approach for time series anomaly detection. This hybrid approach tries to satisfy domain and behavioral variability in vehicle CAN messages by combining a machine learning downstream task with the multi-stage technique, namely Federated Learning (FL). The various domains in the proposed framework are aligned with the various ECUs in a vehicle, which correspond to the federated clients.

The downstream methodology starts with data preprocessing to partition unevenly spaced time series to 2-d fixed-length data frames. Irregularity in the time series segments is captured by creating an auxiliary feature that represents the time interval between consecutive readings. The window length of the frames is a fixed number that corresponds to a specific use case (ex. vehicle CAN messages), and it is highly advised to be tuned per use case for better performance.

The following step is the feature extraction. This approach recommends a Transformer-based deep learning model as a feature extractor. This model is chosen because of its ability to capture temporal positional patterns and characteristics between consecutive readings and compress them into a dense vector. That is, the frame features represent contextual relationships between data points within a sequence in the same way as text or images. Finally, the Transformer outputs a compact 1-d fixed-length numerical representation, namely embedding, ready to be consumed by the next framework's downstream component, XGBoost.

The embeddings generated by the Transformer are used as input features for XGBoost ensemble trees. The Gradient Boosting technique is integrated into an anomaly detection framework by incorporating multiple XGBoost regressors. There is one regressor per available label which tries to predict the number of occurrences of each label inside each frame. The output is a vector of size equal to the number of labels and represents the distributional predictions or distributional counts of different classes (e.g., normal vs. anomaly types) inside a time series frame. This approach allows the system to capture more nuanced information about window's composition and normality status. An alternative simplified approach is to keep a single supervisory target by aggregating readings labels inside a frame with the majority voting method.

The last step of the proposed framework is the Federated Residual Ensembling. Federated clients are created and orchestrated in collaborative training without centralizing raw data. The clients represent different data sources like individual vehicles or ECUs. Each client is iteratively training local models and adds the best performing ones to a global server as ensembles. The final distributional predictions are the summarized predictions of the ensemble models.

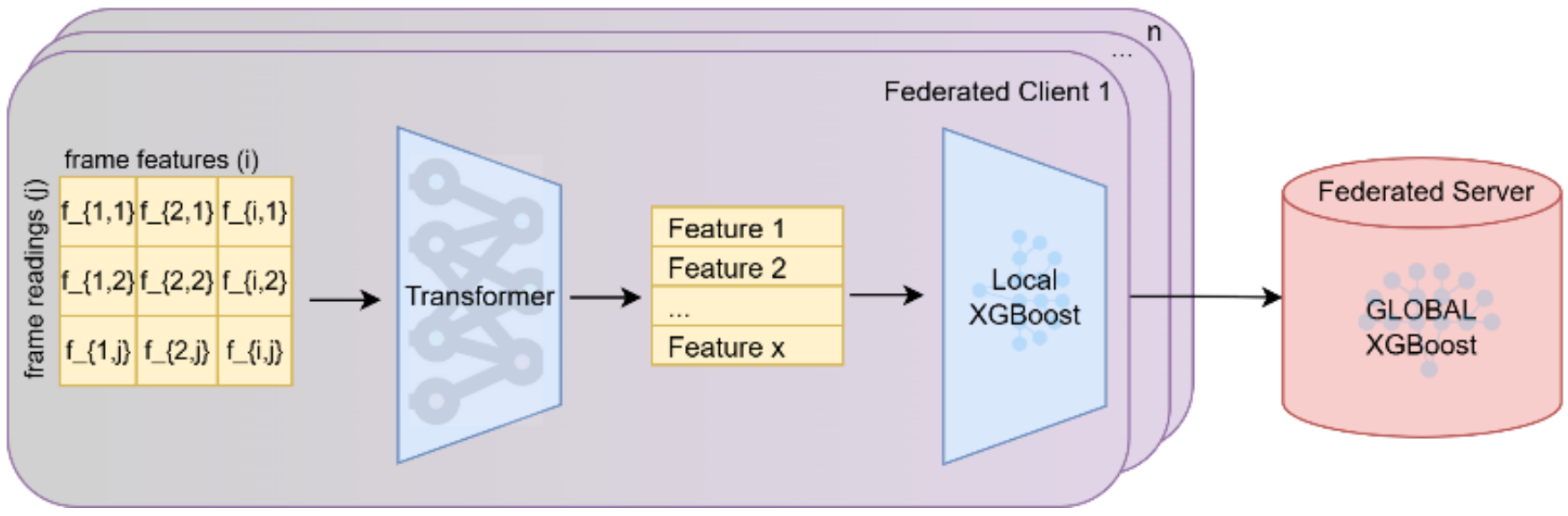

**Figure 1: High-level architecture of the proposed framework**

Overall, the framework balances representational depth, statistical expressiveness, and deployment practicality for anomaly or intrusion detection in structured communication domains like vehicle CAN messages. Figure 1 represents the architecture of the downstream components and the multistage federated clients.

## 4. Dataset Processing & Model Development

### *A. Dataset Overview and Pre-Processing Approach*

The proposed framework was tested on vehicle CAN messages. The data were retrieved from the publicly available Intrusion Dataset provided by the HKSecurity OCSLab [16]. Each CAN message includes a timestamp (TS), an 11-bit identifier (ID1), data length code (DLC) values specifying payload size, and eight payload bytes (DLC0–DLC7). The dataset labels consist of normal type messages and three different attack types (DoS, Fuzzy, impersonation). The CAN messages are emitted by different ECUs making them ideal for ECU based federated learning.

The temporal nature of the data is achieved with several preprocessing steps. Initially, the data are sorted by message identifier and then by timestamp. This sorting preserves the natural flow of the time series and in parallel respects the diversity of each ECU. Regarding feature engineering, eight payload bytes columns and the attack label column are extracted. Also, the time difference between consecutive messages is calculated by ECU. Inter-arrival time feature captures ECU-specific timing patterns and provides the model with temporal cues for detecting anomalies. Finaly, sequence windowing is applied, and the messages are partitioned into non-overlapping segments of size $T$. This windowing strategy preserves the local temporal dependency structure of each ECU while enabling batch-based deep learning-based anomaly detection. The resulting data frames have a shape of $15 \times 9$ $(T \times d_{\text{in}})$, representing $T = 15$ consecutive messages and $d_{\text{in}} = 9$ input features (e.g., 8 payload bytes + Delta_t = 9).

### *B. Temporal Embedding Model for Feature Extraction*

The resulting data frames are converted to an embedding space of dimension $d_{\text{model}} = 64$ by a lightweight Transformer-based encoder. The outcome is high-quality latent representations that preserve both short-range payload structure and long-range temporal dynamics. The encoder has the following architecture:

- a linear input projection layer,
- a learnable positional embedding matrix,
- a stack of $L = 2$ Transformer encoder layers with multi-head self-attention (batch-first configuration),
- an adaptive mean-pooling mechanism producing a pooled embedding for each window.

Formally, for an input sequence $X \in \mathbb{R}^{B \times T \times d_{\text{in}}}$, the encoder computes

$$Z = \text{TransformerEncoder}(XW + P),$$

where $W$ is the input projection matrix and $P$ is the positional embedding. The encoder outputs per-frame embeddings $Z \in \mathbb{R}^{B \times T \times d_{\text{model}}}$ which are averaged across the $T = 15$ messages and finally converted to a dense vector of 64 features per window.

### *C. Multi-Output XGBoost*

The final component to this downstream task is a multi-output regressor based on XGBoost. This regressor is trained on dense vectors of 64 features to predict the distribution of anomaly labels observed within each CAN message window. For every window, the preprocessing stage produces a vector

$$\mathbf{y} = [p_0, p_1, \dots, p_{C-1}],$$

where $p_c$ denotes the empirical proportion of class $c$ within the window. In our case, we have 4 classes and therefore the multi-output wrapper consists of 4 XGBoost regressors with a squared-error loss function. This setup provides a flexible and interpretable mapping from temporal embeddings to anomaly probability distributions. By predicting a full distribution rather than a single label, the system achieves finer granularity in detecting partially anomalous windows and supports rich downstream anomaly-scoring strategies.

## 5. Federated Learning Design and Implementation

### *A. Objective of Federated Learning Integration*

The federated learning framework enables collaborative training of an anomaly detection model across multiple data silos while keeping raw CAN traffic local. This design addresses two main goals:

1. Privacy preservation: Individual clients retain full control over their in-vehicle data, which may include sensitive timing patterns or ECU-specific behaviors.
2. Robust generalization: Training on distributed, heterogeneous data allows the global model to capture diverse ECU behaviors and attack patterns.

In addition, the approach leverages temporal windows of CAN messages rather than individual frames. This design choice reduces computational overhead during inference, making it feasible to deploy the model on edge devices with limited resources.

### *B. Client-Side Dataset Creation*

A specific subset of the CAN OTIDS dataset, CAN_HCRL_OTIDS_UB, was selected exclusively for federated experiments. This subset was never included in the centralized/global model training, ensuring a fair evaluation of federated generalization.

Dataset preparation followed these steps:

1. Temporal window construction: CAN messages were grouped into fixed-size windows per ECU. Each window encapsulates a short sequence of messages, allowing the model to learn temporal patterns while reducing inference cost at deployment.
2. Window-level embedding: Each window is represented as a fixed-size vector capturing both payload values and inter-message timing, producing a compact yet informative representation for classification.
3. Client distribution: Windows were partitioned across multiple client silos so that each client received an exclusive set of ecus. This simulates a realistic decentralized scenario where organizations hold non-overlapping portions of in-vehicle data.
4. Local train/validation/test splits: Each client independently stratified its windows into training, validation, and test sets according to majority-class labels.

This process ensures no single CAN message is processed individually, dramatically reducing inference time, while retaining the sequential structure needed for accurate anomaly detection.

### *C. Local Model Training Logic*

In the federated setting, each client trains a multi-output gradient-boosted model on window-level embeddings derived from its local ECU data. The training strategy is incremental, leveraging the current global ensemble as a starting point. Initially, the client computes predictions on its local training windows using the ensemble, which provides an estimate of the model's current performance. Residual errors are then calculated by comparing these predictions to the true labels. A new gradient-boosted component is trained specifically to fit these residuals, thereby improving the ensemble's representation of local temporal patterns without modifying the previously learned components.

Once the new component is trained, the client evaluates the updated ensemble against its validation set to measure performance improvement. Only the newly trained model component and its validation metrics are communicated back to the central server. This additive procedure allows federated boosting to proceed without averaging model weights, maintaining the integrity of tree-based models while enabling the ensemble to gradually incorporate knowledge from distributed clients. By focusing on window-level embeddings rather than individual CAN messages, this approach also reduces computational load, making it suitable for real-time inference on edge devices.

### *D. Federated Server and Aggregation Strategy*

A central server orchestrates the aggregation of client updates and the evolution of the global model. At the start of each federated round, the server distributes the current global ensemble to all clients, who then perform local training on their window-level embeddings. Clients return their updated model components along with validation metrics, which the server evaluates to identify the most effective contributions. Since tree-based gradient-boosted models cannot be averaged in the traditional sense, the server implements a validation-based selection strategy, sequentially incorporating the best-performing components into the global ensemble.

The server continuously monitors the validation performance across clients and rounds, assessing convergence and ensuring that knowledge from heterogeneous local datasets is effectively integrated. After the completion of all federated rounds, the global model is individually evaluated on each client's test windows to confirm generalization. By combining a select-best aggregation strategy with compact window-level feature representations, the server preserves inference efficiency while enabling collaborative learning, allowing the global model to benefit from diverse ECU behaviors without requiring access to raw CAN traffic.

### *E. Execution Flow*

The general federated learning flow occurs within each client and is coordinated with the global model. The process proceeds as follows:

1. Window selection and feature extraction: Each client selects temporal windows of CAN signals and computes fixed-size embeddings that capture both payload and timing information.
2. Local model update: Using the extracted window-level features, the client trains its local gradient-boosted model on residuals relative to the current global ensemble.
3. Interaction with the global model: The client sends its updated model component and validation metrics to the central server, which incorporates the new component into the global ensemble.
4. Iterative rounds: This exchange of local updates and global ensemble updates continues across multiple federated rounds, allowing the global model to gradually integrate knowledge from all clients.

5. Final evaluation:

   After completion of all rounds, each client evaluates the final global ensemble on its own test windows.

By operating on windows of CAN signals rather than individual frames, this approach significantly reduces computational cost and inference time, making it well-suited for real-time deployment on edge devices.

## 6. Evaluation

Evaluation is organized into centralized (global) model performance and federated model performance, highlighting classification accuracy, regression quality, and efficiency. All results are reported at the window-level, reflecting the temporal embedding approach that reduces inference time on edge devices.

### *A. Centralized Global Model*

The centralized global model was trained on the full embedding dataset prior to federated partitioning and serves as a baseline for comparing federated learning performance. Evaluation considers both classification, where the model predicts the presence and type of attacks, and regression, where it estimates the probability distribution of attack types across temporal windows. Table 1 presents a comprehensive overview of the global model's performance across train, validation, and test splits, showing overall metrics such as accuracy, macro- and weighted precision, recall, and F1-score, along with the mean squared error (MSE) for regression. The table allows direct comparison of performance across splits, illustrating both the model's generalization capability and its ability to maintain consistency on unseen data.

Figure 2 illustrates the per-class error distribution on the test set, where error is defined as one minus the recall for each class. This visualization highlights which attack types are more challenging for the model and provides insight into class-specific detection performance. By focusing on window-level embeddings rather than individual CAN frames, the model efficiently captures temporal patterns while reducing computational overhead, ensuring timely predictions suitable for edge deployment.

The high recall and F1 values for critical attack classes indicate that the model reliably detects anomalies across multiple attack types, while the regression MSE demonstrates that the estimated attack probabilities are accurate. The use of window-level embeddings effectively reduces inference cost, allowing for real-time deployment on edge devices, and the per-class error distribution plot emphasizes where the model may require further improvement or fine-tuning for specific attack categories.

**Table 1: Global Model Metrics (Train / Validation / Test)**

| | train | validation | test |
|---|---|---|---|
| **Accuracy** | 0.859 | 0.834 | 0.838 |
| **Precision (macro)** | 0.874 | 0.85 | 0.852 |
| **Recall (macro)** | 0.86 | 0.837 | 0.84 |
| **F1 (macro)** | 0.861 | 0.837 | 0.84 |
| **Precision (weighted)** | 0.866 | 0.841 | 0.844 |
| **Recall (weighted)** | 0.859 | 0.834 | 0.838 |
| **F1 (weighted)** | 0.856 | 0.831 | 0.835 |
| **Regression MSE** | 0.017 | 0.021 | 0.021 |

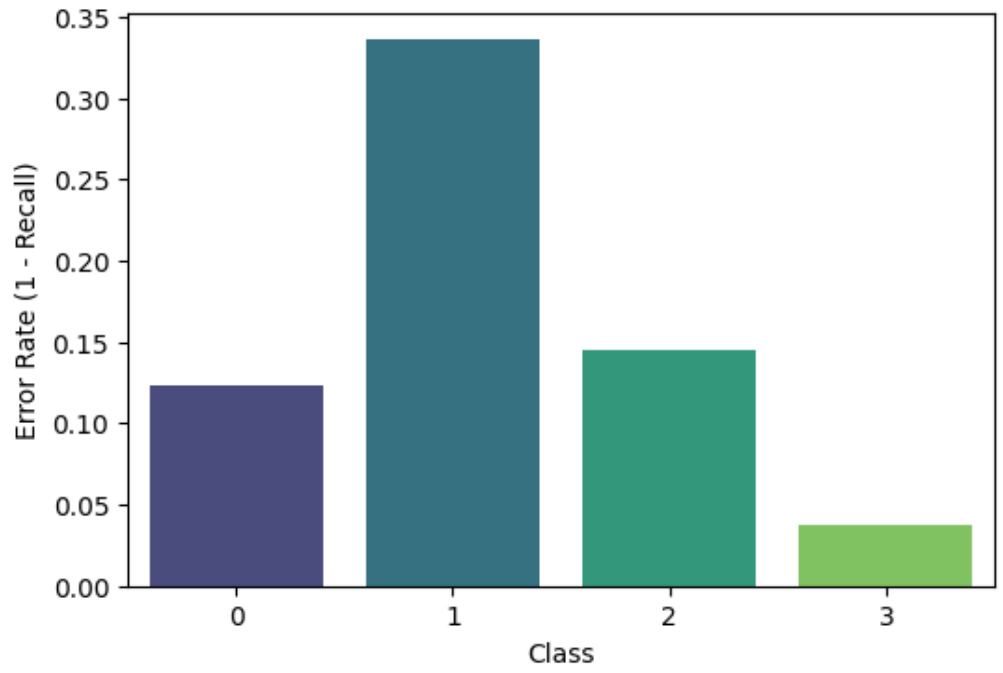

**Figure 2: Per-Class Error Distribution (Test Set)**

***B. Federated Model***

The federated model was trained across two simulated clients, each holding a distinct subset of ECUs from the CAN_HCRL_OTIDS_UB dataset. This setup emulates a realistic decentralized environment where data cannot be shared directly, allowing collaborative model training while maintaining data privacy. Each client computes window-level embeddings locally and trains a residual XGBoost component that is integrated sequentially into the global ensemble.

Validation Metrics Across Federated Rounds

Table 2 summarizes the average validation MSE for each federated round, highlighting the incremental improvement of the global ensemble as knowledge from both clients is incorporated. Figure 3 visualizes the per-client validation MSE across rounds, showing convergence trends and relative contributions of each client to the global model.

**Table 2: Per-Round Validation Metrics of the global ensemble model**

| Round | Average Validation MSE |
|---|---|
| 1.0 | 0.0351 |
| 2.0 | 0.0263 |
| 3.0 | 0.0244 |

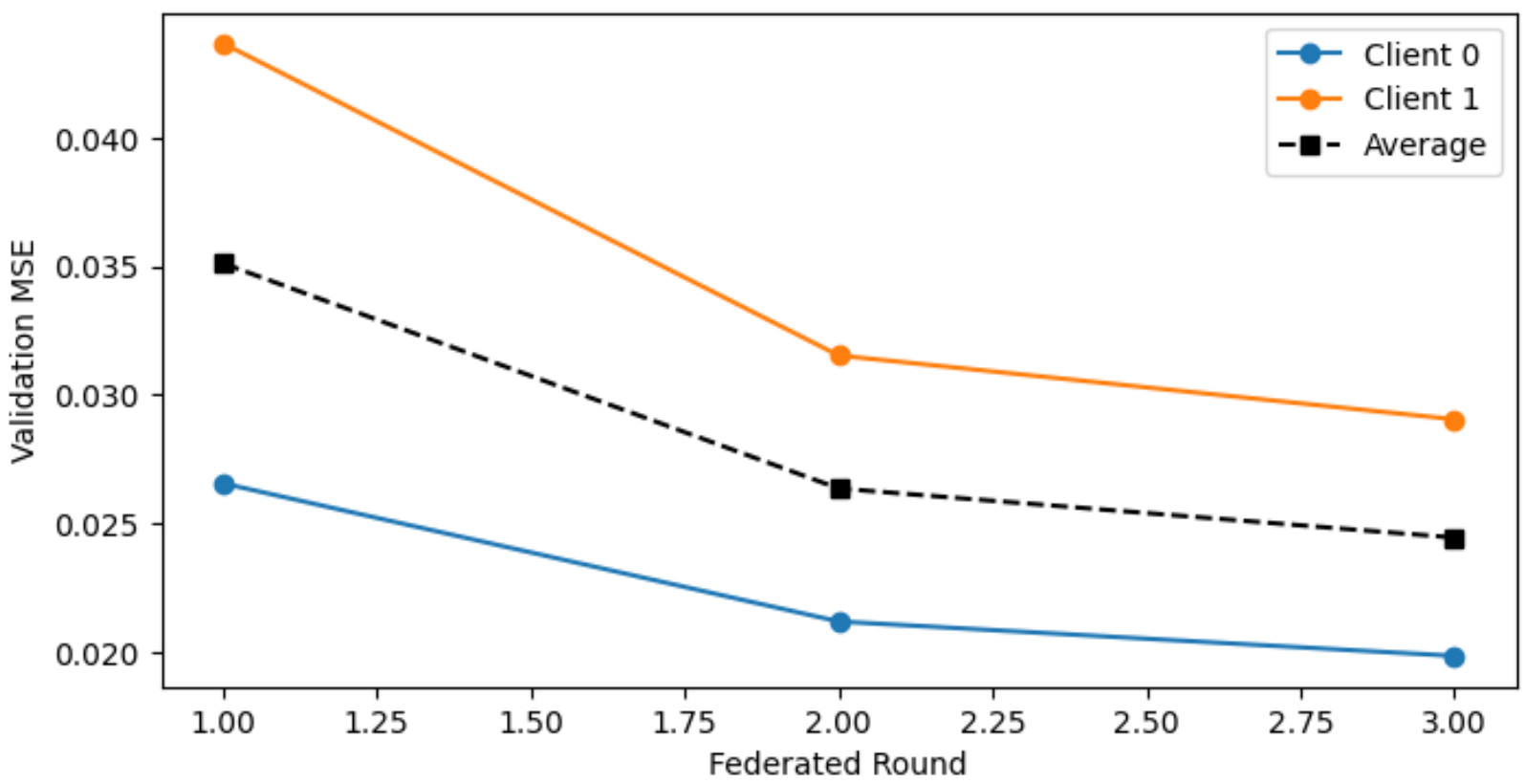

**Figure 3: Validation MSE Across Federated Rounds**

Final Test Metrics per Client

After completion of all federated rounds, the final global ensemble was evaluated on each client's test windows. Table 3 reports the test accuracy and regression MSE for each client, illustrating the model's ability to generalize to unseen ECU-specific data. Figure 4 shows the evolution of test MSE for each client across federated rounds. It illustrates how the incremental integration of knowledge from all clients gradually improves prediction accuracy on unseen test windows, while also highlighting differences in client-specific performance due to heterogeneous local datasets.

**Table 3: Final Test Metrics per Client**

| Client | Test Accuracy | Test MSE |
|---|---|---|
| client 0 | 0.88 | 0.1309 |
| client 1 | 0.89 | 0.1316 |

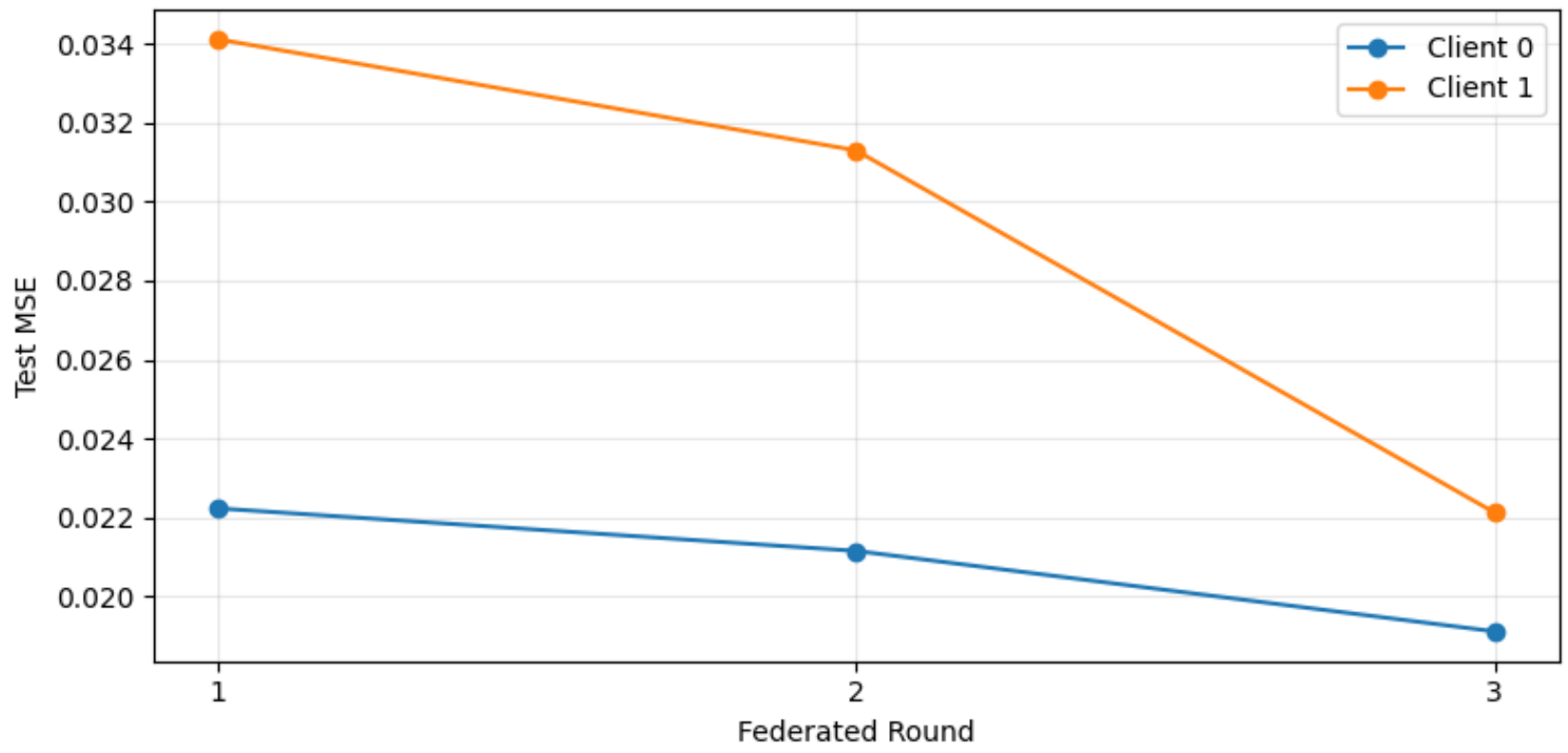

**Figure 4: Test MSE per client across 3 rounds**

The federated model achieves high accuracy and low regression error for both clients, demonstrating effective knowledge integration despite data being siloed. Window-level embeddings contribute to efficient inference, maintaining low computational cost while preserving sequential temporal patterns necessary for robust anomaly detection. Differences in client performance reflect dataset heterogeneity and emphasize the importance of collaborative learning to improve generalization across diverse ECU behaviors.

## 7. Conclusions & Future Work

This paper introduces an anomaly-detection framework for in-vehicle networks relying on CAN protocol. The solution combines a lightweight Temporal Transformer encoder with an efficient XGBoost-based anomaly classifier and a privacy-preserving FL scheme. The inclusion of a lightweight downstream classifier ensures that the system remains suitable for resource-constrained ECUs, while the FL component enables collaborative model improvement without exposing raw CAN data. Future work involves the integration of online learning, so that the model is updated gradually over time, bringing its applicability even closer to real and highly dynamic environments with actual evolving conditions.

**Acknowledgment**

This work has been funded by the Shift2SDV project (Grant Agreement No. 101194245), supported by Chips Joint undertaking and its members, including top-up funding by the national authorities of Austria, Denmark, Germany, Greece, Finland, Italy, Netherlands, Poland, Portugal, Spain, Turkey.